\documentclass[letterpaper, 10 pt, conference]{ieeeconf}  % Comment this line out if you need a4paper

\IEEEoverridecommandlockouts                              % This command is only needed if 
\usepackage{graphicx}
\usepackage{subfig}
\usepackage{epsfig} % for postscript graphics files
\usepackage{mathptmx} % assumes new font selection scheme installed
\usepackage{times} % assumes new font selection scheme installed
\usepackage{amsmath} % assumes amsmath package installed
\usepackage{amssymb}  % assumes amsmath package installed
\usepackage{array}
\usepackage{booktabs}
\usepackage{multirow}
\usepackage{balance}
\usepackage{xspace}
\usepackage{color}

\def \YOLO{Y\textsc{olo}\xspace}
\def \RANSAC{R\textsc{ansac}\xspace}
\title{\LARGE \bf Stable Tracking of Eye Gaze Direction During Ophthalmic Surgery
}

\author{Tinghe Hong, Shenlin Cai, Boyang Li, Kai Huang*% <-this % stops a space
\thanks{The authors are with the school of computer science and engineering, Sun Yat-sen University,Guangzhou, China.}
\thanks{This research is supported by the Shenzhen Medical Research Foundation (Grant No. SZMRF-D2404009) and the Science and Technology Development Fund of Macau Special Administrative Region (Grant No. 0010/2023/AKP).
}%
}

\begin{document}

\maketitle
\thispagestyle{empty}
\pagestyle{empty}

%%%%%%%%%%%%%%%%%%%%%%%%%%%%%%%%%%%%%%%%%%%%%%%%%%%%%%%%%%%%%%%%%%%%%%%%%%%%%%%%
\begin{abstract}

Ophthalmic surgical robots offer superior stability and precision by reducing the natural hand tremors of human surgeons, enabling delicate operations in confined surgical spaces. Despite the advancements in developing vision- and force-based control methods for surgical robots, preoperative navigation remains heavily reliant on manual operation, limiting the consistency and increasing the uncertainty. Existing eye gaze estimation techniques in the surgery, whether traditional or deep learning-based, face challenges including dependence on additional sensors, occlusion issues in surgical environments, and the requirement for facial detection. To address these limitations, this study proposes an innovative eye localization and tracking method that combines machine learning with traditional algorithms, eliminating the requirements of landmarks and maintaining stable iris detection and gaze estimation under varying lighting and shadow conditions. Extensive real-world experiment results show that our proposed method has an average estimation error of 0.58 degrees for eye orientation estimation and 2.08-degree average control error for the robotic arm's movement based on the calculated orientation.

\end{abstract}

%%%%%%%%%%%%%%%%%%%%%%%%%%%%%%%%%%%%%%%%%%%%%%%%%%%%%%%%%%%%%%%%%%%%%%%%%%%%%%%%
\section{Introduction}

Ophthalmic surgical robots demonstrate significant stability and precision compared to human surgeons when performing surgical tasks. The design of robotic arms effectively eliminates or substantially reduces natural hand tremors during surgery, enabling more refined and controlled operations in extremely confined surgical spaces, such as those encountered in retinal surgery. This stability advantage not only increases the success rate of surgeries but also minimizes the risk of damage to surrounding healthy tissues.

Significant progress has been made in the development of ophthalmic surgical robots. Various studies have proposed vision- and force-based compliant control methods for ophthalmic surgical robots to achieve precise movement control of surgical instruments during contact with ocular tissues \cite{WangZSZS23,FanWLYZZ23}. However, despite the advancements in surgical robotics, the preoperative navigation in ophthalmic surgery largely remains dependent on manual operation. This reliance not only limits the consistency and repeatability of surgeries but also increases the uncertainty during the surgical process.

In the field of ophthalmic surgery, preoperative and intraoperative navigation serve distinct roles. Particularly before surgical operations commence, the lack of physical contact between the instruments and ocular tissues prevents surgical robots from utilizing force feedback mechanisms for accurate position determination \cite{he2014multi}. Consequently, preoperative navigation primarily relies on eye-gaze estimation techniques guided by visual cues to achieve precise positioning of the robotic arm. Currently, eye-gaze estimation techniques are categorized into two main types: traditional edge detection-based algorithms \cite{kar2017review} and the more recent deep learning-based approaches \cite{ghosh2023automatic}, \cite{cheng2024appearance}.

\begin{figure}[hpbt]
	\centering
	\includegraphics[width=0.95\linewidth]{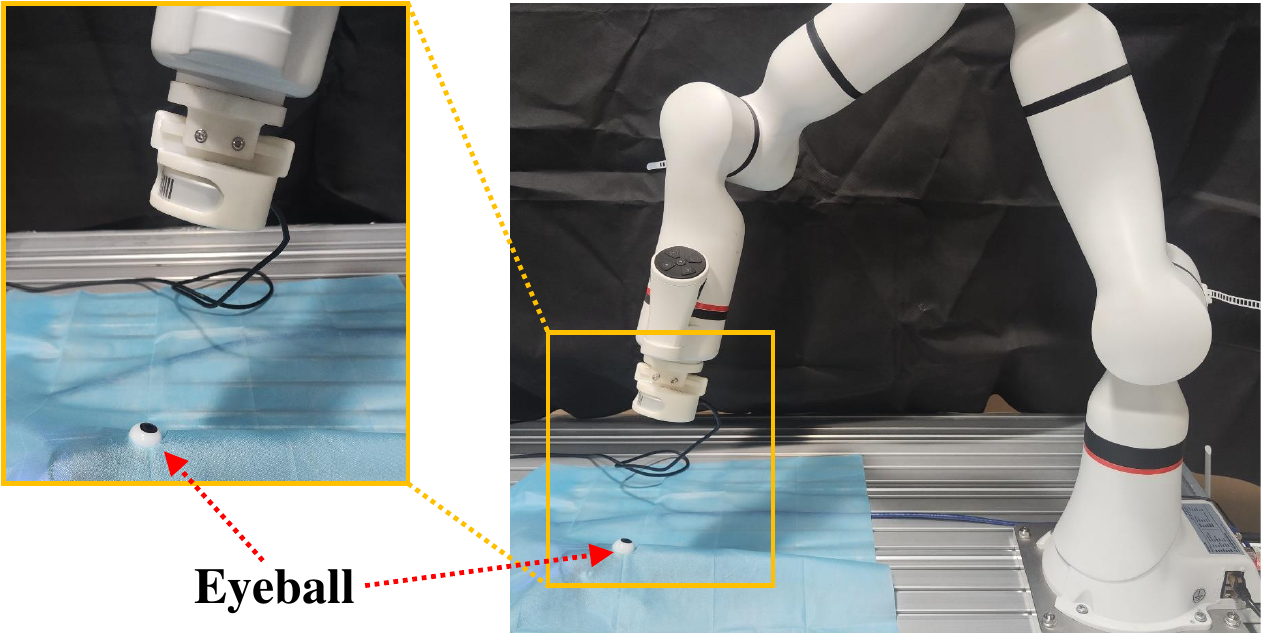}
	\caption{A mock ophthalmic surgical environment with a robotic arm.}
	\label{fig:sence}
\end{figure}

There are several key challenges in recognizing eye gaze direction in surgical environments (see Fig. \ref{fig:sence}). Firstly, the primary perception equipment used in the surgery is a variable focal length endoscope. \cite{ModelE11} and \cite{LiL14} utilized Kinect depth sensors for real-time eye tracking. Although the methods are simple and low-cost \cite{ZhouCSYL16}, they requires the introduction of depth-sensing devices, thereby increasing the uncertainty in the surgical setting. Secondly, some methods choose to identify the eyes using fixed-point light reflections \cite{MeyerBMB06}, \cite{MorimotoAF02}. Unfortunately, maintaining stable light reflection positions in the operating room is challenging, as they are often obstructed during surgery. Lastly, the patient's face are typically covered by a surgical pad sheet during the surgery, making the face not visible. Some learning-based methods \cite{0001HDK22}, \cite{Marin-JimenezKM22} are less sensitive to environmental interference. For instance, the authors in \cite{ChengB022} employed an unsupervised approach to improve algorithm performance across different domains, while \cite{kothari2021weakly} used a weakly supervised method for unconstrained gaze estimation. These methods generally require initial face detection before identifying the eye's position and orientation, lacking stability during the tracking process.

To address the aforementioned challenges, this study proposes an eye localization and tracking method that integrates machine learning techniques with traditional algorithms. Firstly, a fine-tuned \YOLO \cite{redmon2016you} model is employed to accurately identify the sclera and iris regions, effectively overcoming the limitations associated with face detection. Then, the \RANSAC method is used to stably determine the eye position within the identified region, enhancing the robustness of the tracking process. Finally, the positional relationship between the eye and pupil is used to resolve ambiguities in gaze estimation. Our approach provides a stable solution for eye localization and tracking suitable for use with robotic arms. Extensive real-world experiments demonstrate that the proposed method achieves an average orientation estimation error of $0.58^\circ$, an average robotic arm control error of $2.08^\circ$, and a relative distance error of 6.4mm between the eye and camera.

\section{Related Works}

In research on eye gaze tracking, accurate eye localization is a fundamental and critical step, involving the recognition of the iris or sclera. In \cite{yiu2019deepvog}, researchers estimate the center position of the eye by calculating the intersections of multiple lines passing through the edges of the iris. However, this method does not address the problem of precise localization of the iris center in the visual frame. Another study \cite{canu2019eye} proposed an approach based on facial feature point detection, where key facial landmarks are firstly identified. The eye-related features are further recognized. While this method can assist in eye localization to some extent, it remains limitations by the recognizability of facial features.

In estimating eye gaze direction, the method proposed in \cite{schroff2015facenet} requires obtaining facial orientation information before accurately estimating the eye's gaze, which is challenging in cases of facial occlusion or unclear vision. Similarly, the approaches in \cite{catruna2024crossgaze} and \cite{Park2019ICCV} rely on accurate recognition of facial features for eye gaze estimation, leading to the reduced accuracy or failure to estimate when facial features are indistinct or obscured. Additionally, the method in \cite{chaudhary2019ritnet} requires facial features such as eyelids to estimate eye gaze direction, thus also being limited by the detectability of the features. Although the study in \cite{yiu2019deepvog} provides gaze information under certain conditions, its effectiveness depends on the relative stability of the camera and eye positions, which is difficult to maintain in a dynamically changing surgical environment. In \cite{dehghani2022colibridoc}, the authors achieve surgical robot navigation by identifying trocars, providing an innovative solution for eye gaze estimation. However, this method requires pre-placement of trocars on the sclera.

\section{Method}

The proposed method in this study consists of several steps: (1) accurately and stably identifying the position of the iris in the camera frame under varying lighting conditions; (2) calculating the eye gaze direction using the camera's intrinsic parameters and the position of the iris; (3) resolving the ambiguity in gaze estimation by analyzing the relative positions of the iris and sclera centers.

\subsection{Iris Position Acquisition} To determine the position of the iris in an image, a fine-tuned \YOLO model is first employed to identify the region containing the iris, avoiding the target loss issues caused by interference from other similar objects in the camera's field of view, which can occur when directly using the pupiltrack method mentioned in \cite{Nithin}. The \YOLO model is trained on a dataset of approximately one hundred annotated images of eyes. The fine-tuned model demonstrates robustness against interference and can accurately detect the general region of the iris, although its determination of the precise iris position remains somewhat unstable.

To achieve stable localization of the iris in the image, the region identified is firstly subjected to a binarization process to distinguish the darker and the lighter area of the sclera.

To enhance the stability of detection and ensure consistent differentiation of the iris's dark region under varying shadow or lighting conditions, an adaptive thresholding method for binarization is employed.

\begin{equation}
thd = avg*k.
\end{equation}

\begin{figure}[hpbt]
	\centering
	\begin{minipage}[t]{0.31\linewidth}
		\centering
		\subfloat[b][Original. ]{\label{fig:original}\includegraphics[width=\textwidth]{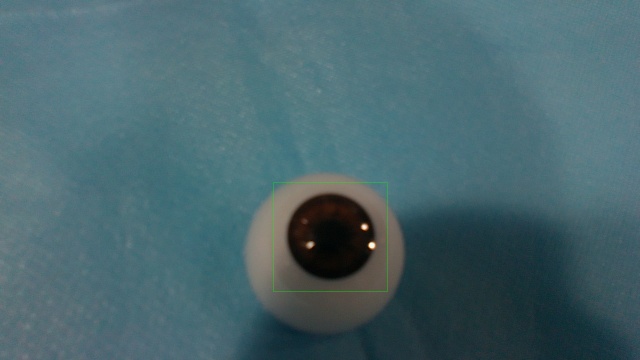}} %'image-a'
	\end{minipage}
	\begin{minipage}[t]{0.31\linewidth}
		\centering
		\subfloat[b][Fixed. ]{\label{fig:fixed_threshold}\includegraphics[width=\textwidth]{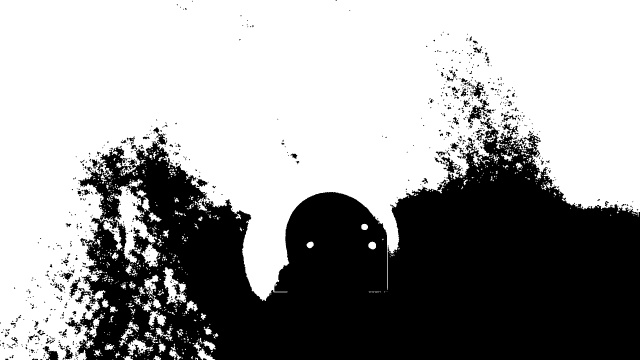}} %'image-b'
	\end{minipage}
	\begin{minipage}[t]{0.31\linewidth}
		\centering
		\subfloat[b][Adaptive. ]{\label{fig:adaptive_threshold}\includegraphics[width=\textwidth]{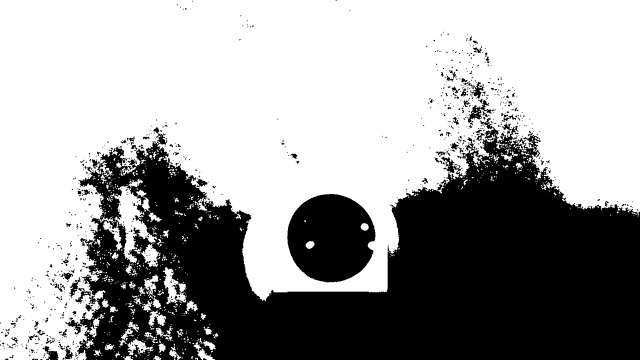}} %'image-b'
	\end{minipage}
	\caption{Binarization results with fixed and adaptive threshold under shadow conditions.} 
\end{figure}

Here, $thd$ represents the threshold, $avg$ is the mean color value within the region, $k$ is a scaling factor. When the overall brightness within the region decreases, the threshold will also decrease accordingly to ensure that the iris region can still be distinguished after binarization (see Fig. \ref{fig:fixed_threshold} and Fig. \ref{fig:adaptive_threshold}). Additionally, performing the assessment within the defined region helps to avoid interference from variations in ambient lighting and shadows.

%Subsequently, edge detection is performed within the defined region (Fig. \ref{fig:contour}), followed by the extraction of the convex hull (Fig. \ref{fig:hull}). Because the region is predefined, it is not affected by interference from other objects. The convex hull with the largest area is then selected as the boundary surrounding the iris, and the points along the edge of this convex hull are chosen (Fig.\ref{fig:chosen_hull}). Finally, an ellipse is fitted to the selected points, representing the position of the iris in the image (Fig.\ref{fig:ellipse}).

\begin{figure}[hpbt]
	\vspace{7pt}
	\centering
	\begin{minipage}[t]{0.4\linewidth}
		\centering
		\subfloat[b][Edge detection. ]{\label{fig:contour}\includegraphics[width=\textwidth]{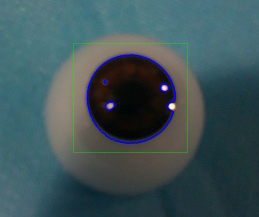}} %'image-a'
	\end{minipage}
	\hspace{5pt}
	\begin{minipage}[t]{0.4\linewidth}
		\centering
		\subfloat[b][Convex hull. ]{\label{fig:hull}\includegraphics[width=\textwidth]{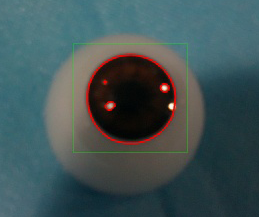}} %'image-b'
	\end{minipage}

	\begin{minipage}[t]{0.4\linewidth}
		\centering
		\subfloat[b][\RANSAC iteration. ]{\label{fig:chosen_hull}\includegraphics[width=\textwidth]{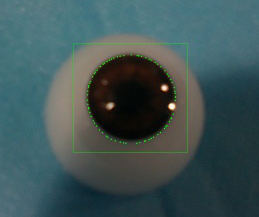}} 
	\end{minipage}
	\hspace{5pt}
	\begin{minipage}[t]{0.4\linewidth}
	\centering
	\subfloat[b][Fit ellipse. ]{\label{fig:ellipse}\includegraphics[width=\textwidth]{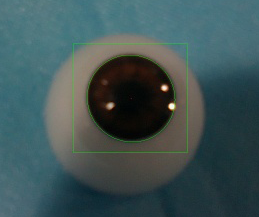}} 
\end{minipage}
	\caption{Binarization results with different thresholds under shadow conditions.} 
\end{figure}

Within the defined region, edge detection is firstly performed to extract points located at the boundaries (Fig. \ref{fig:contour}). It can be observed that some reflective spots on the iris are also identified as edge points, which introduces a certain degree of interference in the recognition process. Subsequently, a convex hull is constructed based on the detected edge points (Fig. \ref{fig:hull}). This step helps to filter out the interfering points located at the edges. However, some noise points within the iris region are also detected as a separate convex hull. To accurately identify the true convex hull corresponding to the iris, we utilize the region defined by \YOLO and select the largest convex hull by area, which represents the iris, to effectively filter out smaller convex hulls formed by noise points. The vertices of this convex hull are then extracted (Fig. \ref{fig:chosen_hull}). Finally, an ellipse is fitted to the selected vertices, providing the precise position of the iris (Fig. \ref{fig:ellipse}).

\subsection{Calculating Eye Position and Gaze Direction}

\begin{figure}[hpbt]
	\centering
	\includegraphics[width=0.85\linewidth]{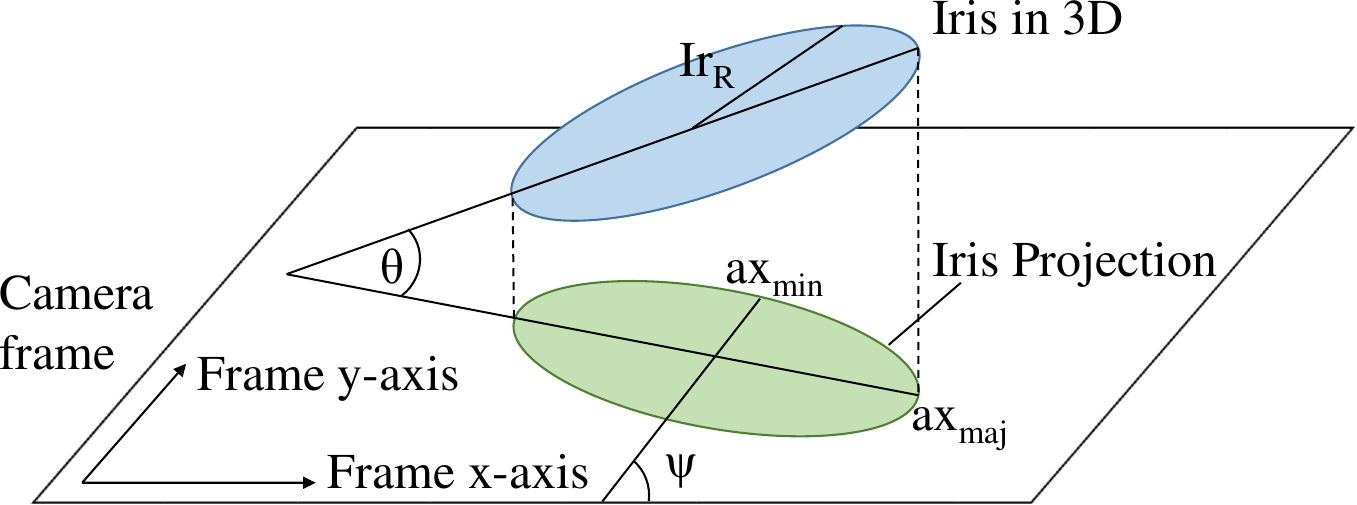}
	\caption{The projection of the iris from 3D space onto the 2D camera frame.}
	\label{fig:pos}
\end{figure}

The ellipse fitted in the previous section represents the projection of the iris in 3D space onto the camera frame (see Fig. \ref{fig:pos}), The iris in 3D space can be considered as a circle with a radius of $Ir_R$, measured in millimeters. 

Let the major and minor axes of the ellipse defined as $ax_{maj}$ and $ax_{min}$ (in pixel units), respectively. The center of the ellipse in the image is located at $p_x$ and $p_y$, with the ellipse's rotation angle relative to the image's x-axis being $\psi$. $f_x$ and $f_y$ represent the x-axis and y-axis pixel focal lengths of the camera, while $pr_x$ and $pr_y$ denote the pixel coordinates of the screen's center. To compute the coordinates of the eyeball in three-dimensional space, let $f_z=(f_x+f_y)/2$. Based on the similar triangles, the relative coordinates $[Ir_x, Ir_y, Ir_z]$ of the iris to the camera frame center are given by:

\begin{equation}
\left\{
\begin{aligned}
Ir_z &= f_z*Ir_R/ax_{maj} \\
Ir_x&=-Ir_z*(p_x-pr_x)/f_x\\
Ir_y&=Ir_z*(p_y-pr_y)/f_y. \!
\end{aligned}
\right.
\end{equation}

%Here, $Ir_R$ represents the radius of the iris, measured in millimeters.
The rotation angle $\theta$ of the eye relative to the camera plane is defined as the follows:
\begin{equation}
\theta=\arccos(ax_{min}/ax_{maj}).
\end{equation}

Thus, the normal vector $\vec{n}$ of the eye gaze direction is defined as follows:

\begin{equation}
\vec{n}=[-\sin(\theta)*\cos(\psi),-\sin(\theta)*\sin(\psi), \cos(\theta)]^{\top}.
\end{equation}

When the eye is not centered in the camera's field of view, the obtained eye gaze direction needs to be adjusted. Firstly, the distance $d$ between the center of the ellipse and the center of the field of view is calculated by:
\begin{equation}d = (Ir_{x}^{2}+Ir_{y}^{2})^{1/2}.\end{equation}
The angle of rotation $\gamma$ required is given by,
\begin{equation}\gamma=\arctan(d/Ir_z).\end{equation}

The rotation axis $\vec{l}$ is given by,
\begin{equation}\vec{l}=[Ir_y/d,Ir_x/d,0]^{\top}.\end{equation}

The normal vector $\vec{n}$ of the eye gaze direction is corrected using Rodrigues' rotation formula, resulting in the corrected normal vector $\vec{n'}$,
\begin{equation}\vec{n'}=\vec{n} \cdot \cos(\gamma)+(\vec{l}\times\vec{n})\cdot\sin(\gamma)+\vec{l}(\vec{l} \cdot \vec{n})(1-\cos(\gamma)).\end{equation}

\subsection{Ambiguities in Eye Gaze Direction}

In the process of recovering the 3D eye orientation from 2D images, ambiguity arises because a single projection may correspond to two different eye orientations. Therefore, resolving this ambiguity is essential to accurately determine the eye's orientation.

\begin{figure}[hpbt]
	\centering
	\includegraphics[width=0.75\linewidth]{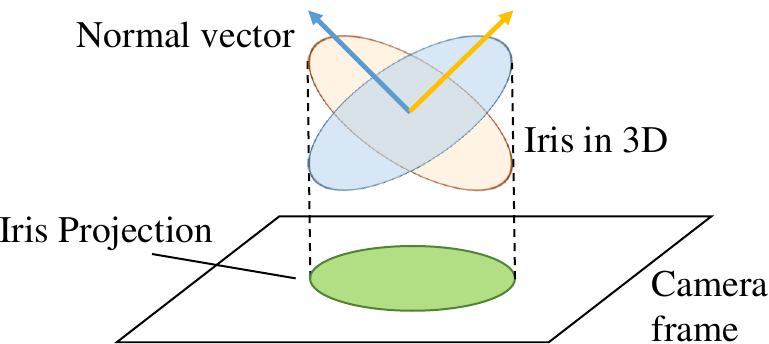}
	\caption{For an elliptical projection, two possible circular orientations can be reconstructed.}
	\label{fig:amb}
\end{figure}

\begin{figure}[hpbt]
	\centering
	\includegraphics[width=0.7\linewidth]{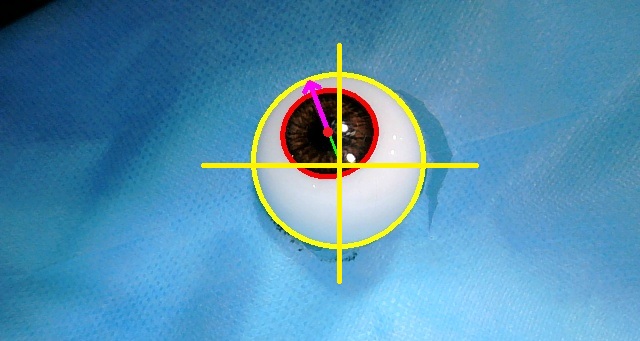}
	\caption{The true orientation of the eye is determined by comparing the relative positions of the iris and sclera centers.}
	\label{fig:arrow}
\end{figure}

As shown in Fig. \ref{fig:amb}, when the 2D projection in the camera is reconstructed into the 3D space of the iris, two distinct solutions can arise, corresponding to the two different normal vectors. To resolve this, a method similar to that described in Section 3.A is used to obtain the sclera's pixel coordinates, which are then compared with the iris's pixel coordinates to determine the true orientation of the iris (see Fig. \ref{fig:arrow}).

\section{Experiments}

To validate the stability and accuracy of the proposed method, three sets of experiments are designed. The first experiment compares the iris recognition performance. The second experiment evaluates the accuracy of eye-gaze direction estimation and the ability to resolve ambiguities. Finally, in the third experiment, a camera is mounted on a robotic arm. The proposed method is used to compute the eye's position and orientation, enabling the end-effector of the robotic arm to follow the eye's movements in real-time. \textcolor{black}{The proposed approach runs on a computer equipped with an Intel i5-10400 processor and an NVIDIA Titan V GPU for YOLO inference.}

\subsection{2D Iris Recognition}

In this experiment, the camera remains stationary while the eye model rotates around a fixed axis. The goal of the experiment is to compare the different methods for iris recognition and tracking in the camera's field of view. Five different methods are evaluated: our proposed method, the \RANSAC-based Pupiltrack \cite{Nithin}, the learning-based GazeML method \cite{park2018learning}, the CNN-based Pupil-Locator \cite{eivazi2019improving}(Abbreviated as Locator), and the 3D Eye Tracker \cite{plopski2016automated}.

%The experiment was conducted under two different resolutions. The first setup had a resolution of 1920 $\times$ 1080 at a frequency of 30 Hz, with the camera positioned approximately 100 mm from the eye. Under these conditions, the 3D Eye Tracker failed to detect the iris. Consequently, a second experiment was performed with a resolution of 640 $\times$ 360 at 30 Hz, with the camera at a distance of about 50 mm from the eye. However, under these conditions, the Locator failed to detect the iris. The eye rotated for a duration of 10 seconds, during which 300 frames were captured. Table 1 shows the number of frames where each method lost tracking out of the 300 frames. Cases where detection was not possible are indicated by the "-" symbol in the data table. The instances of frame loss for each recognition method are shown in Table \ref{tab:lost_frame}.

The experiments are conducted under two different resolution settings. The first setting uses a resolution of 1920 $\times$ 1080 at 30Hz, with the camera positioned approximately 100mm from the eye. However, under these conditions, 3D Eye Tracker failed to detect the iris. To address this, the camera is adjusted to a position approximately 50mm from the eye, with the resolution set to 640 $\times$ 360 at 30Hz. At this closer distance, Locator struggled to detect the iris. Additionally, GazeML required landmarks to identify the iris, \YOLO is employed to predefine the region containing the eye for more effective detection.

We set the eyeball rotation duration to 10 seconds, rotating around an axis parallel to the y-axis of the camera frame, from $-30^\circ$ to $+30^\circ$, resulting in a total of 300 frames for analysis. A tracking failure is defined as a deviation exceeding 30 pixels. Tab. \ref{tab:lost_frame} presents the number of frames in which each recognition method lost tracking out of the 300 frames. Cases where recognition was not possible are denoted by a "-" symbol. The frame loss statistics for each method are displayed in Tab. \ref{tab:lost_frame}.

\begin{table}[hpbt]
	\centering
    \caption{The number of lost frames in iris detection tasks in 300-frame videos at two different resolutions.}
	\begin{tabular}{lll}
		\toprule
					   & Resolution 1920$\times$1080 & Resolution 640$\times$360\\ 
					   \midrule
		Ours           & 0         & 0       \\ 
		GazeML\cite{park2018learning}         & 3         & 16      \\ 
		Pupiltrack\cite{Nithin}     & 5         & 10      \\ 
		Locator\cite{eivazi2019improving}        & -         & 0       \\ 
		3D Eye Tracker\cite{plopski2016automated} & 7         & -       \\ 
		\bottomrule
	\end{tabular}
	\label{tab:lost_frame}
\end{table}

As shown in Tab. \ref{tab:lost_frame}, all methods except ours experienced varying degrees of tracking loss. By using \YOLO to define the region of the eye and employing an adaptive thresholding approach, our method is able to consistently detect the iris's position in the image under various lighting conditions.

\begin{figure}[hpbt]
	\centering
	\begin{minipage}[t]{0.4\linewidth}
		\centering
		\subfloat[b][GazeML. ]{\label{fig:GazeML}\includegraphics[width=\textwidth]{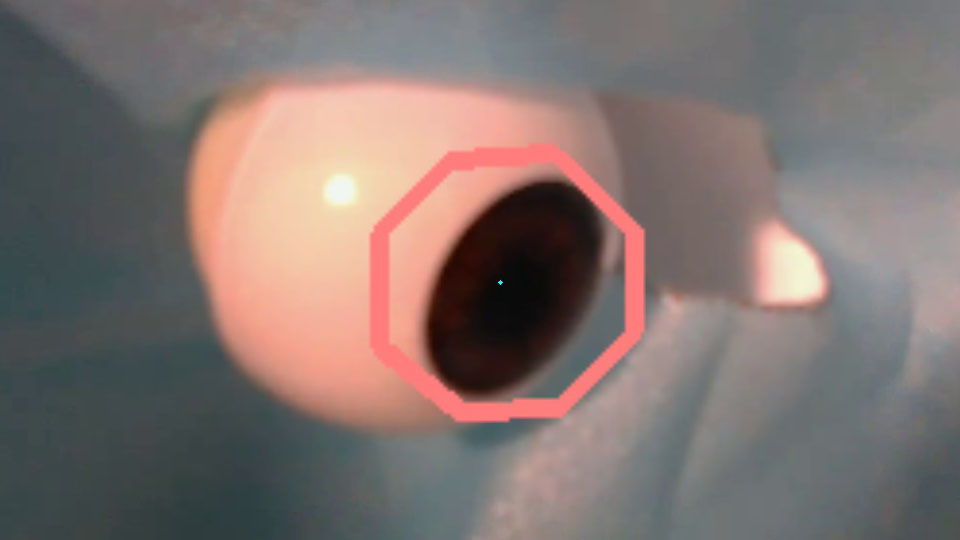}} %'image-b'
	\end{minipage}
	\begin{minipage}[t]{0.4\linewidth}
		\centering
		\subfloat[b][Pupiltrack. ]{\label{fig:pupil_tracke}\includegraphics[width=\textwidth]{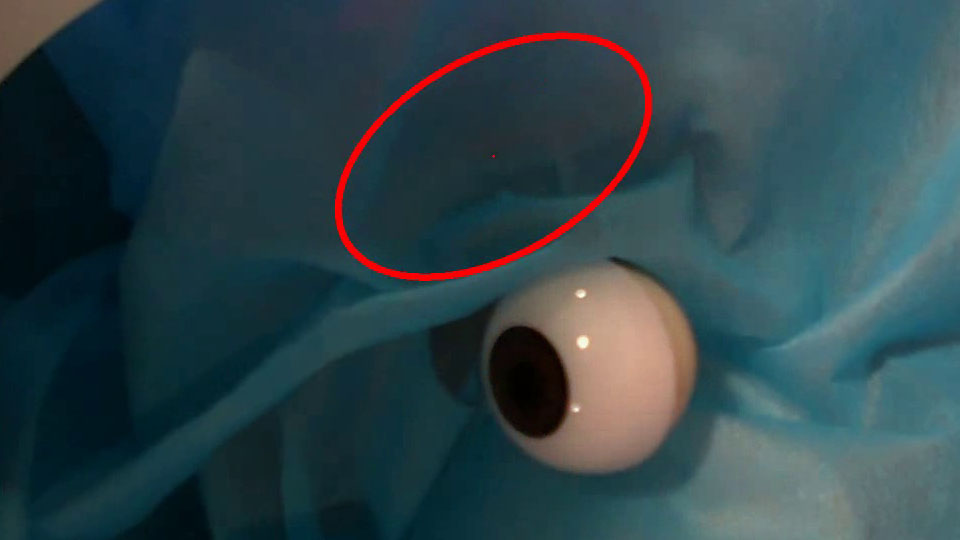}} %'image-a'
	\end{minipage}

	\begin{minipage}[t]{0.4\linewidth}
		\centering
		\subfloat[b][Locator. ]{\label{fig:locator}\includegraphics[width=\textwidth]{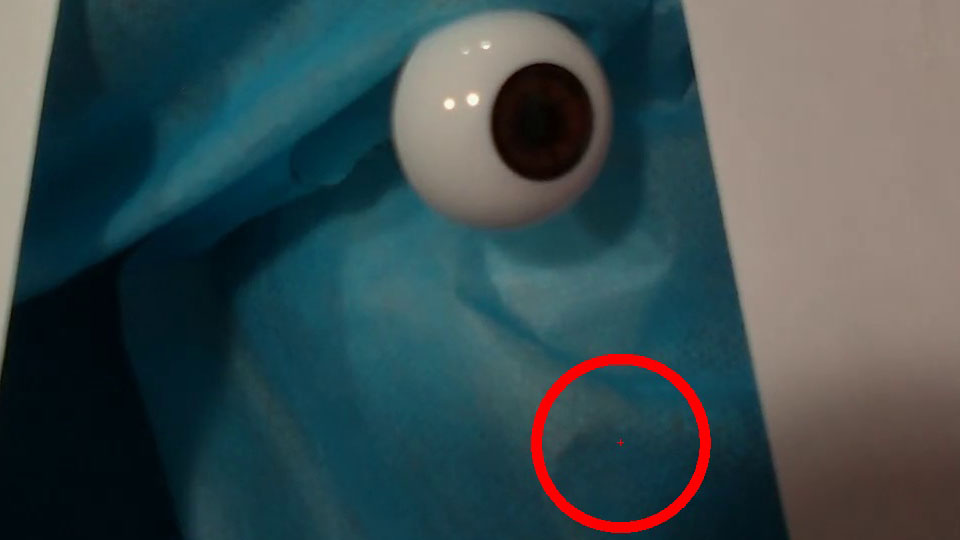}} 
	\end{minipage}
	\begin{minipage}[t]{0.4\linewidth}
		\centering
		\subfloat[b][3D Eye Tracker. ]{\label{fig:eye_tracker}\includegraphics[width=\textwidth]{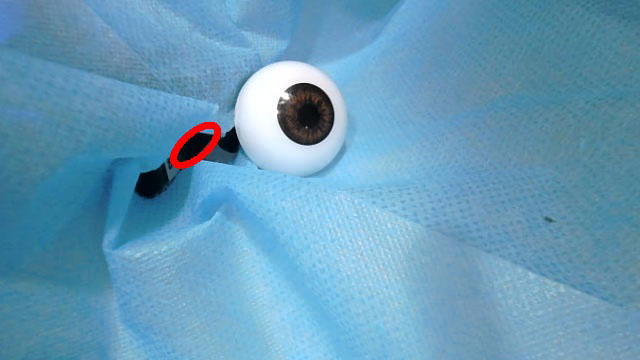}} 
	\end{minipage}
	\caption{The instances of frame loss for each method.} 
	\label{fig:loss}
\end{figure}

Fig. \ref{fig:loss} illustrates the tracking failures of other methods under conditions with lighting variations and shadows. The Locator (Fig. \ref{fig:locator}) performs well when the camera is positioned at a greater distance from the eye (Resolution 1920$\times$1080), but exhibits significant tracking loss at closer distances (Resolution 640$\times$360). Both Pupiltrack (Fig. \ref{fig:pupil_tracke}) and 3D Eye Tracker(Fig. \ref{fig:eye_tracker}) are highly susceptible to interference from shadows. The GazeML (Fig. \ref{fig:GazeML}) tends to deviate in recognition process, leading to unstable positioning.

\begin{figure*}[hpbt]
	\vspace{10pt}
	\centering
	\begin{minipage}[t]{0.45\linewidth}
		\centering
		\subfloat[b][Iris center at resolution 1920$\times$1080. ]{\label{fig:near}\includegraphics[width=\textwidth]{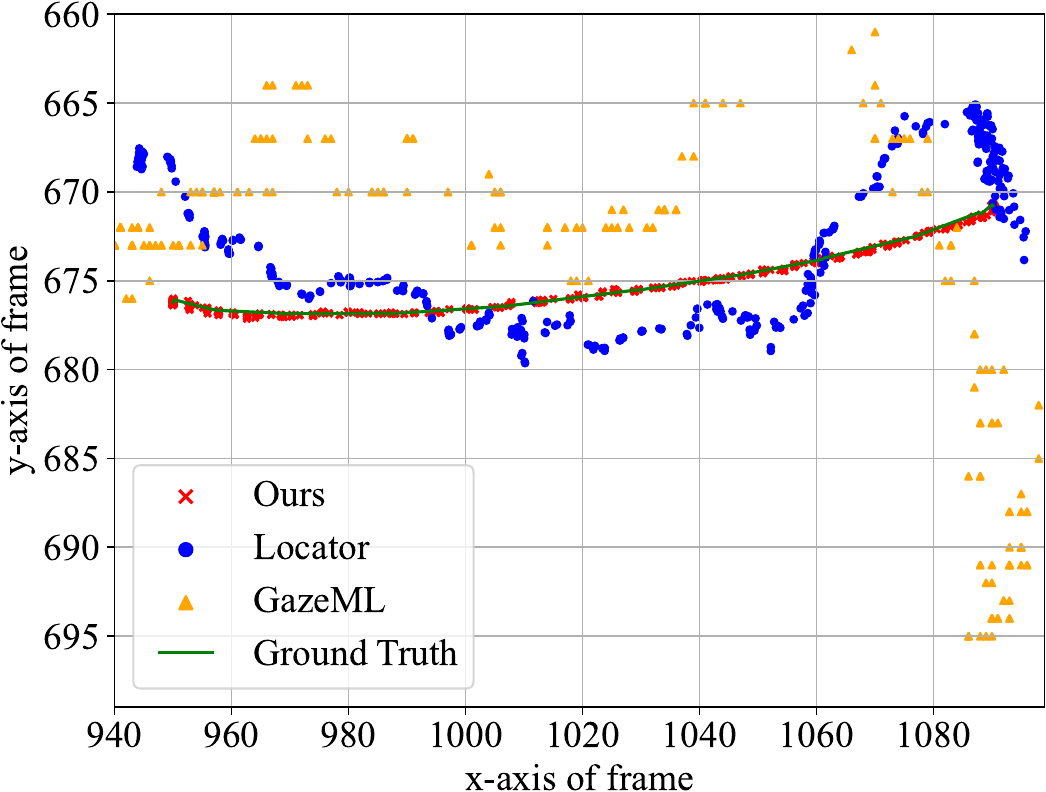}} %'image-b'
	\end{minipage}
	\hspace{10pt}
	\begin{minipage}[t]{0.45\linewidth}
		\centering
		\subfloat[b][Iris center at resolution 640$\times$360. ]{\label{fig:far}\includegraphics[width=\textwidth]{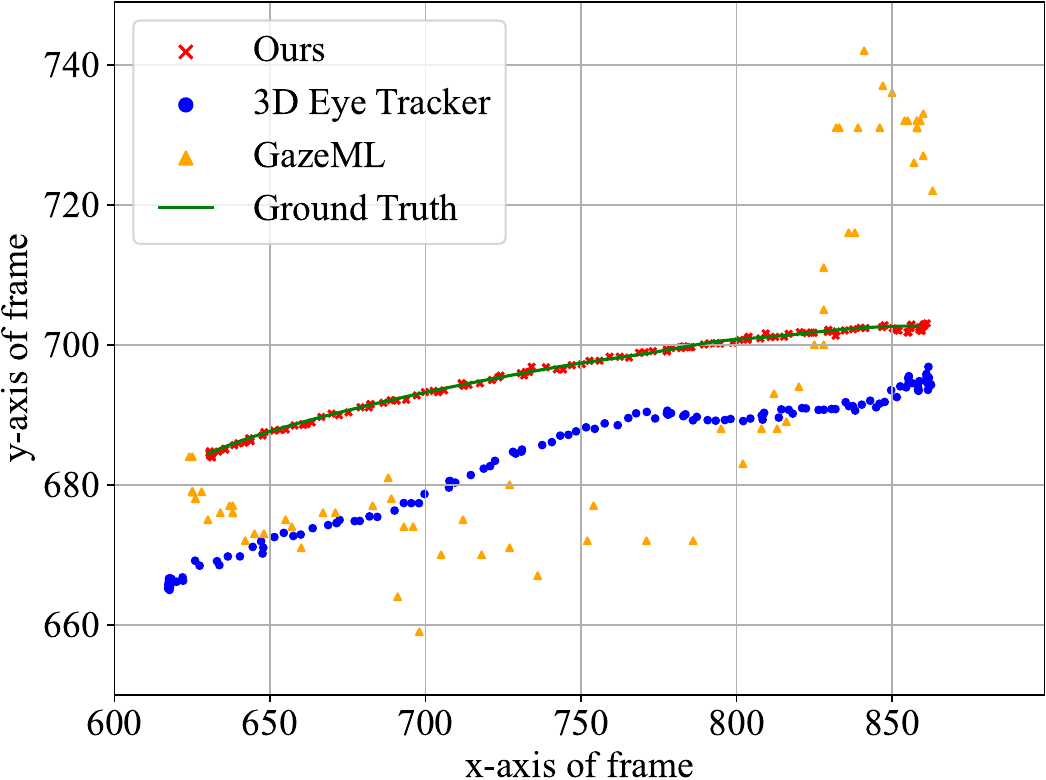}} %'image-a'
	\end{minipage}
	\caption{The instances of frame loss for each recognition method.} 
	\label{fig:loss-f}
\end{figure*}

Fig. \ref{fig:loss-f} depicts the positional shifts of various recognition techniques within the camera's coordinate system throughout the eye movement. The horizontal and vertical axes represent pixel measurements. For clarity, only the coordinates of points where the tracking remains uninterrupted are shown. It should be noted that Pupiltrack is omitted from the figure due to its coordinates closely aligning with those of our proposed method when tracking is uninterrupted.

As shown in Fig. \ref{fig:loss-f}, our method closely matches the Ground Truth during the tracking process. In contrast, the two learning-based methods (GazeML and Locator) demonstrate less stability. The detailed tracking results are presented in Tab. \ref{tab:loss-v}.

Tab. \ref{tab:loss-v} presents the mean and standard deviation of pixel deviation from the ground truth for different methods at two resolutions. The mean metric represents the accuracy of each tracking method, while the standard deviation metric indicates the stability of the tracking. Since points with significant deviation (considered as tracking loss) are excluded from the statistics, our method and the Pupiltrack method show similar performance. The GazeML method exhibits a higher level of jitter, while the Locator and 3D Eye Tracker methods are relatively more stable but exhibit lower adaptability.

\begin{table}[hpbt]
	\centering
    \caption{The mean and standard deviation of the error from the ground truth during tracking, measured in pixels.}
	\begin{tabular}{ccccc}%
		\toprule%
		&\multicolumn{2}{c}{Resolution 1920$\times$1080}&\multicolumn{2}{c}{Resolution 640$\times$360}\\
		&Mean&Standard&Mean&Standard\\
		\midrule
		Ours&0.119&0.103&0.749&0.278 \\
		\midrule
		GazeML&8.631&6.591&17.019&10.439 \\
		\midrule
		Pupiltrack&0.095&0.087&0.814&0.258 \\
		\midrule 
		Locator&3.423&2.421&-&- \\
		\midrule
		3d Eye Tracker&-&-&11.704&3.835 \\
		\bottomrule	
	\end{tabular}
	\label{tab:loss-v}
\end{table}

\subsection{3D Eyeball Orientation Estimation}

In this experiment, we firstly set the eyeball to rotate 30 degrees around an axis parallel to the x-axis of the camera frame. Then, we control the eyeball to rotate from $-30^\circ$ to $+30^\circ$ around an axis parallel to the y-axis of the camera frame while calculating the normal vector of the eyeball relative to the camera.

In this experiment, we do not include a comparison with Locator, as it cannot provide a 3D eyeball orientation. Additionally, the 3D Eye Tracker requires a pre-established model to compute the orientation, which is not suitable for our current experimental setup.

\begin{figure}[hpbt]
	\centering
	\begin{minipage}[t]{0.47\linewidth}
		\centering
		\subfloat[b][GazeML. ]{\label{fig:GazeML_3D}\includegraphics[width=\textwidth]{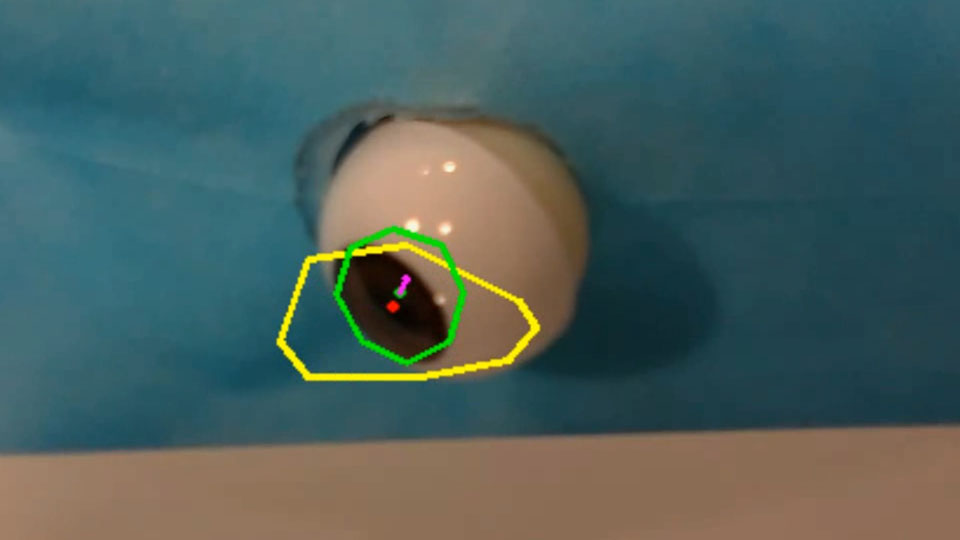}} %'image-b'
	\end{minipage}
	\begin{minipage}[t]{0.47\linewidth}
		\centering
		\subfloat[b][Pupiltrack. ]{\label{fig:Pupiltrack_3D}\includegraphics[width=\textwidth]{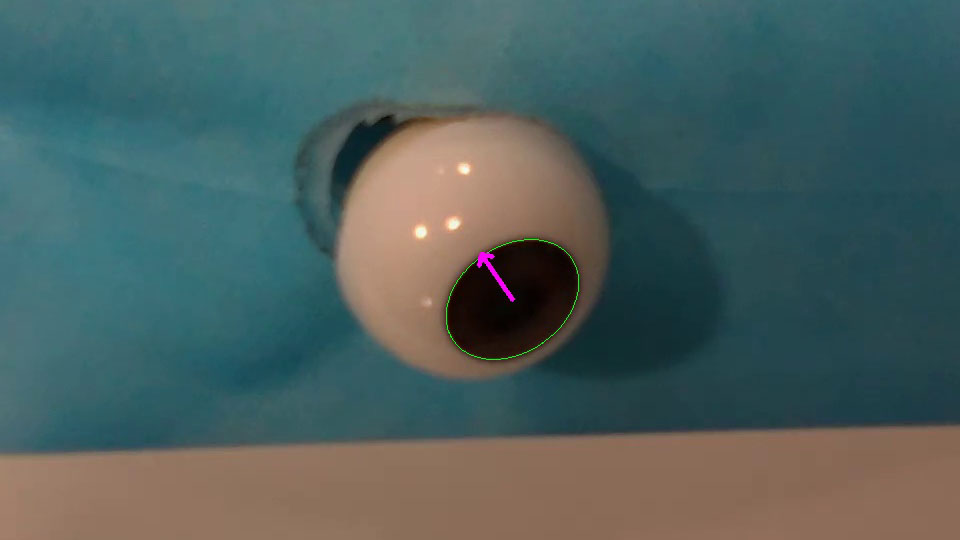}} %'image-a'
	\end{minipage}
	\caption{Errors in eye orientation estimation.} 
	\label{fig:3D_error}
\end{figure}

The purple arrows in Fig. \ref{fig:3D_error} represent the eye orientations estimated by different algorithms. The GazeML method suffers from instability when detecting the iris position, which adversely affects the accuracy of its eye orientation estimation (see Fig. \ref{fig:GazeML_3D}). As a result, the estimated normal vector significantly deviates when the eye deflection angle is large. On the other hand, Pupiltrack lacks a mechanism to resolve ambiguity, leading to instances where the direction is incorrectly identified as the opposite (see Fig. \ref{fig:Pupiltrack_3D}).

\begin{figure}[hpbt]
	\centering
	\begin{minipage}[t]{0.47\linewidth}
		\centering
		\subfloat[b][Orientation in 2D.
		 ]{\label{fig:3D_real}\includegraphics[width=\textwidth]{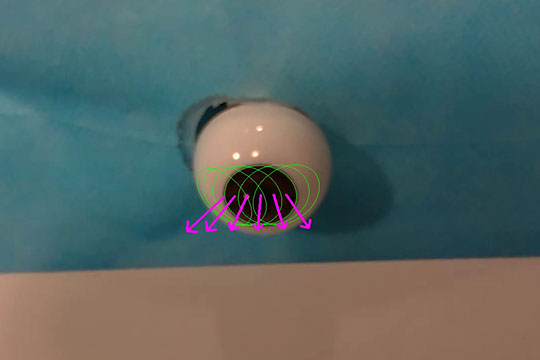}} %'image-b'
	\end{minipage}
	\begin{minipage}[t]{0.47\linewidth}
		\centering
		\subfloat[b][Orientation in 3D. ]{\label{fig:3D_cod}\includegraphics[width=\textwidth]{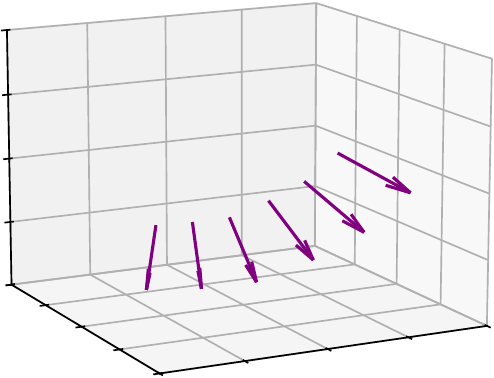}} %'image-a'
	\end{minipage}
	\caption{Eyeball position and orientation in 2D image and 3D coordinate system.} 
	\label{fig:3D}
\end{figure}

Profit from stable tracking, our method also demonstrates consistent performance in estimating eye orientation. Furthermore, by resolving the ambiguity issue, our approach avoids misjudgment of the eye's direction. Even when the significant changes occur in the projection's rotation angle ($\pi s$), the method accurately identifies the eye's orientation throughout the rotation process. The experimental results are illustrated in Fig. \ref{fig:3D}.

Fig. \ref{fig:3D} illustrates the estimated eye orientation and position using our proposed method. The purple arrows indicate the estimated eye orientation. Fig. \ref{fig:3D_real} shows the estimated orientation and position of the eye plotted in the camera frame while Fig. \ref{fig:3D_cod} displays the estimated eye position and orientation in a 3D coordinate system. From the results, it can be concluded that our method exhibits both stability and accuracy in estimating the eye's position and orientation.

In Tab. \ref{tab:estimation}, we present the estimated normal vectors and their corresponding ground truth values at 5-degree intervals. Angle metric denotes the rotation angle of the eye, Estimation metric represents the estimated normal vector. Ground Truth metric indicates the actual normal vector of the eye. Error metric shows the angular deviation.

\begin{table}[hpbt]
	\vspace{10pt}
	\centering
    \caption{Comparison of estimated normal vectors and ground truth normal vectors at various eye rotation angles, along with the corresponding angular errors.}
	\begin{tabular}{cccccccc}%
		\toprule%
		Angle & \multicolumn{3}{c}{Estimation(x,y,z)} & \multicolumn{3}{c}{Ground truth(x,y,z)} & Error\\
		\midrule
$-30^\circ$  & 0.428  & 0.514 & -0.742 & 0.433  & 0.5 & -0.750 & $0.968^\circ$ \\
$-25^\circ$  & 0.364  & 0.507 & -0.780 & 0.365  & 0.5 & -0.784 & $0.514^\circ$ \\
$-20^\circ$  & 0.299  & 0.501 & -0.811 & 0.296  & 0.5 & -0.813 & $0.303^\circ$ \\
$-15^\circ$  & 0.227  & 0.494 & -0.838 & 0.224  & 0.5 & -0.836 & $0.418^\circ$ \\
$-10^\circ $ & 0.150  & 0.491 & -0.857 & 0.150  & 0.5 & -0.852 & $0.572^\circ$ \\
$-5^\circ$   & 0.073  & 0.490 & -0.868 & 0.075  & 0.5 & -0.862 & $0.659^\circ$ \\
$0^\circ$    & 0.006  & 0.494 & -0.869 & 0.0    & 0.5 & -0.866 & $0.517^\circ$ \\
$5^\circ$    & -0.076 & 0.493 & -0.866 & -0.075 & 0.5 & -0.862 & $0.406^\circ$ \\
$10^\circ$   & -0.150 & 0.493 & -0.856 & -0.150 & 0.5 & -0.852 & $0.451^\circ$ \\
$15^\circ$   & -0.224 & 0.497 & -0.838 & -0.224 & 0.5 & -0.836 & $0.188^\circ$ \\
$20^\circ $  & -0.292 & 0.505 & -0.811 & -0.296 & 0.5 & -0.813 & $0.351^\circ$ \\
$25^\circ$   & -0.365 & 0.515 & -0.775 & -0.365 & 0.5 & -0.784 & $1.010^\circ$ \\
$30^\circ$   & -0.429 & 0.518 & -0.739 & -0.433 & 0.5 & -0.750 & $1.241^\circ$ \\
		\bottomrule	
	\end{tabular}
	\label{tab:estimation}
\end{table}

As shown in Tab. \ref{tab:estimation}, our method not only resolves the ambiguity issue but also achieves a small deviation between the estimated and true angles, with an average error of $0.584^\circ$. Generally, larger errors occur when the deflection angle approaches to $\pm30^\circ$. This can be attributed to the fact that, as the deflection angle increases, the ratio between the major and minor axes changes more drastically, thereby amplifying the error.

\subsection{Eye Tracking with a Surgical Robotic Arm}

In this experiment, we mounted a camera on a 7-degrees robotic arm and rotate the eye model to allow the robotic arm to track the eye's movement. This setup is designed to test the stability and accuracy of our proposed method for eye tracking.

\begin{figure}[hpbt]
	\centering
	\includegraphics[width=0.65\linewidth]{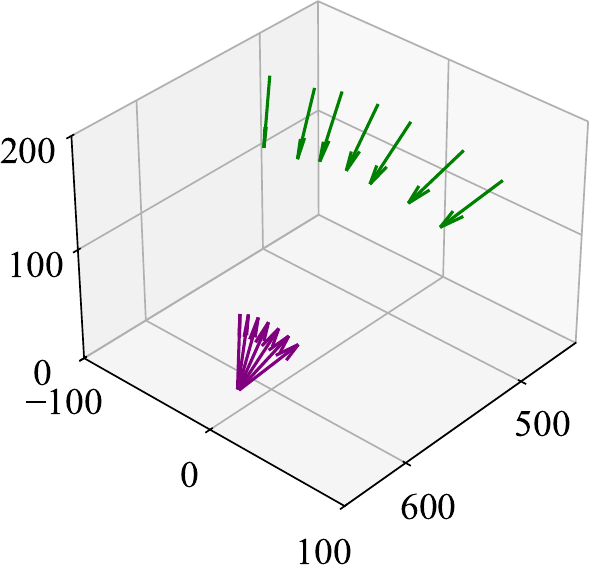}
	\caption{The position and orientation of the eye model and the camera.}
	\label{fig:arrow_robot}
\end{figure}

The experimental setup is shown in Fig. \ref{fig:sence}, where a RealSense camera is mounted at the end of a robotic arm with an eye model rotated by a motor. In the world coordinate system, the eye rotates around the x-axis from -30 degrees to +30 degrees. %The robotic arm uses the camera to detect the eye's deviation and follow its rotation accordingly. 
\textcolor{black}{In this experiment, the end-effector of the robotic arm is controlled to perform rotational or translational movements, ensuring that the camera remains oriented directly towards the iris while maintaining a fixed distance from it.} We record the position and orientation of the robotic arm's end effector at 10-degree intervals, compared them with the computed position and orientation of the eye, and plotted the results in Fig. \ref{fig:arrow_robot}, coordinates are in millimeters.

In Fig. \ref{fig:arrow_robot}, the coordinate system is based on the world coordinate frame, with units in millimeters. \textcolor{black}{The purple arrows indicate the orientation of the eyeball and the position of the iris center, which are computed based on the motor rotation values along with the dimensions of the eyeball and iris. The green arrows represent the position and orientation of the camera controlled by the robotic arm}. As the eyeball rotates, the end effector of the robotic arm, guided by our method, also rotates around the eyeball. Throughout the entire process, the camera maintains a consistent orientation facing the eyeball and keeps a relatively fixed distance from it. The specific data is presented in Tab. \ref{tab:arm}.

\begin{table}[hpbt]
	\centering
    \caption{Comparison of estimated normal vectors and ground truth normal vectors at various eye rotation angles, along with the corresponding angular errors.}
	\begin{tabular}{ccc}
		\toprule%
		Eyeball angles& Distance(mm)            &Error angles                \\
		\midrule
		$-30^\circ$ & 212.189   & $3.750^\circ$\\
		$-20^\circ$ & 206.088  & $2.918^\circ$     \\
		$-10^\circ$ & 208.020    & $1.792^\circ$    \\
		$0^\circ$   & 208.424  & $1.187^\circ$    \\
		$10^\circ$  & 207.224   & $0.681^\circ$     \\
		$20^\circ$  & 211.029  & $3.255^\circ$    \\
		$30^\circ$  & 212.557 & $0.967^\circ$\\
		\bottomrule	
	\end{tabular}
	\label{tab:arm}
\end{table}

Tab. \ref{tab:arm} presents the distance and orientation angle errors between the camera and the eye when the eye is rotated to seven different angles. In the table, "eyeball angles" represents the angles of rotation of the eye around the x-axis in the world coordinate system, "distance" indicates the distance between the camera and the eye, and "error angles" refer to the angle between the eye's orientation and the camera's orientation (with the camera orientation inverted to calculate the angle). Throughout the entire process, the average distance between the camera and the eye was 209.362 mm, and the average error angle was $2.078^\circ$.  \textcolor{black}{According to Tab. \ref{tab:estimation}, approximately $1.494^\circ$  of error can be attributed to the position and orientation control of the robotic arm's end-effector.} The table shows that under our method's control, the surgical robotic arm can maintain a consistent distance from the eye and continuously keep the camera aligned with the eye.

\section{Conclusion}
In this study, we propose a robust method for estimating the 3D orientation of the eyeball from 2D images. Our approach effectively addresses the ambiguity in determining the eyeball's orientation by leveraging adaptive thresholding and the accurately detect the iris region under varying lighting conditions. Through extensive experiments, we demonstrate that our method outperforms other existing approaches in both stability and accuracy. Our method shows consistent performance across different resolutions and lighting conditions. Additionally, our approach accurately estimates the eyeball's orientation with minimal errors, particularly when the rotation angles are small, and resolves ambiguity by comparing the relative positions of the iris and sclera centers. Overall, our method offers a significant improvement in tracking and orientation estimation of the eyeball, which could be beneficial for various applications in computer vision and human-computer interaction.

\balance
\bibliographystyle{unsrt}
\bibliography{main}

\end{document}